\documentclass[conference]{IEEEtran}

\usepackage{mathtools,amssymb,amsthm,bbm}
\usepackage{xcolor}%
\usepackage{graphicx}
\usepackage[font=small,labelfont=bf]{caption}
\usepackage{float}
\usepackage{calc}
\usepackage[square,comma,super,sort&compress]{natbib}

\newcommand{\figtopadj}{}
\newcommand{\figbotadj}{}

\DeclareMathOperator\unif{unif}

\theoremstyle{plain}
\newtheorem{thm}{Theorem}
\newtheorem{rem}[thm]{Remark}

\begin{document}

\title{SAI\\a Sensible Artificial Intelligence that plays Go}

\author{\IEEEauthorblockN{1\textsuperscript{st} Francesco Morandin}
	\IEEEauthorblockA{
		\textit{Università di Parma}\\
		francesco.morandin@unipr.it}
	\and
	\IEEEauthorblockN{2\textsuperscript{nd} Gianluca Amato}
	\IEEEauthorblockA{
		\textit{Università di Chieti-Pescara}\\
		gianluca.amato@unich.it}
	\and
	\IEEEauthorblockN{3\textsuperscript{rd} Rosa Gini}
	\IEEEauthorblockA{
		\textit{Agenzia regionale di sanità della Toscana}\\
		rosa.gini@ars.toscana.it}
	\and
	\IEEEauthorblockN{4\textsuperscript{th} Carlo Metta}
	\IEEEauthorblockA{
		\textit{Università di Firenze}\\
		carlo.metta@gmail.com}
	\and
	\IEEEauthorblockN{5\textsuperscript{th} Maurizio Parton}
	\IEEEauthorblockA{
		\textit{Università di Chieti-Pescara}\\
		maurizio.parton@unich.it}
	\and
	\IEEEauthorblockN{6\textsuperscript{th} Gian-Carlo Pascutto}
	\IEEEauthorblockA{
		\textit{Mozilla Corporation}\\
		gcp@sjeng.org}
}

\maketitle

\begin{abstract}
	We propose a multiple-komi modification of the AlphaGo Zero/Leela Zero paradigm. The winrate as a function of the komi is modeled with a two-parameters sigmoid function, hence the winrate for all komi values is obtained, at the price of predicting just one more variable. A second novel feature is that training is based on self-play games that occasionaly branch --with changed komi-- when the position is uneven. With this setting, reinforcement learning is shown to work on 7$\times$7 Go, obtaining very strong playing agents. As a useful byproduct, the sigmoid parameters given by the network allow to estimate the score difference on the board, and to evaluate how much the game is decided. Finally, we introduce a family of agents which target winning moves with a higher score difference.
\end{abstract}

\section{Introduction}

The longstanding challenge in artificial intelligence of playing Go at
professional human level has been successfully tackled in recent
works~\cite{Alphago,AlphaZero,AlphaGoZero}, where software tools
(AlphaGo, AlphaGo Zero, AlphaZero) combining neural networks and Monte
Carlo tree search reached superhuman level. Such techniques can be generalised, see for instance \cite{metamodeling,AlphaSeq,evolution}.
A recent development was
Leela Zero~\cite{LeelaZero}, an open source software whose neural
network is trained over millions of games played in a distributed
fashion, thus allowing improvements within reach of the resources of
the academic community.

However, all these programs suffer from a relevant limitation: it is
impossible to target their margin of victory. They are trained with a
fixed initial bonus for white player (\emph{komi}) of 7.5 and they are built to maximize the winning
probability, without any knowledge of the game score difference.

This has several negative consequences for these programs: when they
are ahead, they choose suboptimal moves, and often win by a small
margin (see many of the games not ending 
in a resignation in \cite{invisible}); they cannot be used with komi 6.5, which is also common in
professional games; they show bad play in handicap games, since the
winrate is not a relevant attribute in that situations.

In principle all these problems could be overcome by replacing the
binary reward (win=1, lose=0) with the game score difference, but the
latter is known to be less robust~\cite{GelAlt2006,SilAlt2007} and in
general strongest programs use the former since the seminal
works~\cite{Coulom2006,GelAlt2006,GelSil2007}.

Truly, letting the score difference be the reward for the AlphaGo Zero
method, where averages of the value are computed over different
positions, would lead to situations in which a low probability of
winning with a huge margin could overcome a high probability of
winning by 0.5 points in MCTS search, resulting in weaker play.

An improvement that would ensure the robustness of estimating winning
probabilities, but at the same time would overcome these limitations,
would be the ability to play with an arbitrary number of bonus points.
The agent would then maximize the winning
probability with a variable virtual bonus/malus, resulting in a
flexible play able to adapt to positions in which it is ahead or
behind taking into account implicit information about the score
difference.
The first attempt in this direction gave unclear
results~\cite{MultipleKomi}.

In this work we propose a model to pursue this strategy, and as a
proof-of-concept  we apply it to
7$\times$7 Go.

The source code of the SAI fork of Leela Zero and of the corresponding
server can be found on GitHub at \url{https://github.com/sai-dev/sai}
and \url{https://github.com/sai-dev/sai-server}.





\section{General ideas}


\subsection{Winrate}
The winrate $\rho$ of the current player depends on the
state $s$. For the sake of generality we include a second parameter,
i.e.~a number $x\in\mathbb Z$ of virtual bonus points for the current player. So we will have $\rho=\rho(s,x)=\rho_s(x)$, with the latter
being our standard notation. When trying to win by some amount of
points $n$, the agent may let $x=-n$ to ponder its chances. 

Since $\rho_s(x)$ as a function of $x$ must be increasing and map the
real line onto $[0,1]$, a family of sigmoid functions is a natural
choice:
\begin{equation}\label{e:rho_def}
\rho_s(x)
  =\sigma(x+\bar k_s,\alpha_s,\beta_s)
\end{equation}
Here we set
\begin{equation}\label{e:sigma_def}
  \sigma(x,\alpha,\beta)
  :=\frac1{1+\exp(-\beta(\alpha+x))} 
\end{equation}
The number $\bar k_s$ is the signed komi, i.e.~if the real komi of the
game is $k$, we set $\bar k_s=k$ if at $s$ the current player is white
and $\bar k_s=-k$ if it is black.

The number $\alpha=\alpha_s$ is a shift parameter: since
$\sigma(-\alpha,\alpha,\beta)=1/2$, it represents the expected
difference of points on the board from the perspective of the current
player.
The number $\beta=\beta_s$ is a scale parameter: the higher it is,
the steeper is the sigmoid, generally meaning that the result is
set.
The highest meaningful value of $\beta$ is of the order of 10,
since at the end of the game, when the score on the board is set,
$\rho$ must go from about 0 to about 1 by increasing its argument by
one single point.
The lowest meaningful value of $\beta$ for the full 19$\times$19 board
is of the order of $10/2/361\approx0.01$, since at the start of the
game, even for a very weak agent it would be impossible to lose with a
361.5 points komi in favor.

\subsection{Neural network: duplicate the head}
\label{s:model_nn}

AlphaGo, AlphaGo Zero, AlphaZero and Leela Zero all share the same core
structure, with neural networks
that for every
state $s$ provide
\begin{itemize}
\item a probability distribution over the possible moves $p_s$ (the
  \emph{policy}), trained as to choose the most promising moves for
  searching the tree of subsequent positions;
\item a real number $v_s$ (the \emph{value}), trained to estimate the probability of winning for the
current player.
\end{itemize}

We propose a modification of Leela Zero neural network that for every
state $s$ gives the usual policy $p_s$, and the two parameters
$\alpha_s$ and $\beta_s$ described above instead of $v_s$.


%

\subsection{Branching from intermediate position}

Training of Go neural networks with multiple komi evaluation is a
challenge on its own.
Supervised approach appears unfeasible, since large databases of games
have typically standard komi values of 6.5, 7.5 or so and moreover
it's not possible to estimate final territory reliably for them.
Unsupervised learning asks for the creation of millions of games even
when the komi value is fixed. If that had to be made variable, then
theoretically millions of games would be needed \emph{for each komi
  value}%
\footnote{%
  The argument that one can play the games to the end and then score
  under multiple komi does not work here because this doesn't allow to
  estimate the $\beta$ parameter. Moreover that approach would rely on
  the agent of the self-plays to converge to \emph{score}-perfect
  play, while the current approach is satisfied with convergence to
  \emph{winning}-perfect play.}.

Moreover, games started with komi very different from the natural
values may well be weird, wrong and useless for training, unless one
is able to provide agents with different strength.
Finally, we are trying to train two parameters $\alpha_s$ and
$\beta_s$ from a single output, i.e.~the game outcome. To this aim, it would be advisable 
to have at least two finished
games, with different komi, for many training states $s$.

We propose a solution to this problem, by dropping the usual choice
that self-play games for training always start from the initial empty
board position.
The proposed procedure is the following.
\begin{enumerate}
\item Start a game from the empty board with random komi close to the
  natural one.
\item For each state in the game, take note of the estimated value of
  $\alpha$.
\item After the game is finished, look for states $s$ in which
  $d:=|\bar k_s+\alpha_s|$ is large: these are positions in
  which one of the sides was estimated to be ahead of $d$ points.
\item With some probability start a new game from states $s_*$ with
  the komi corrected by $d$ points, in such a way that the new game
  starts with even chances of winning, but with a komi very different
  from the natural one.
\item Iterate from the start.
\end{enumerate}
With this approach games branch when they become uneven, generating
fragments of games with natural situations in which a large komi may
be given without compromising the style of game.
Moreover, the starting \emph{fuseki} positions, that, with the typical naive
approach, are greatly over-represented in the training data, are in
this way much less frequent.
Finally, not all but many training states are in fact branching points
for which there exists two games with different komi, 
yielding easier training.

\subsection{Agent behaviour}
\label{s:agent_model}

We incorporated in our agents the following smart choices of Leela Zero: 
%
\begin{itemize}
\item the evaluation of the winrate of an intermediate
  state $s$ is the \emph{average} of the value $v$ over the subtree of
  states rooted at $s$, instead of the typical \emph{minimax} that is
  expected in these situations;
\item the final selection of the move to play is done, at the root of
  the MCTS tree, by maximizing the number of playouts instead of the
  winrate.
\end{itemize}
 
However, we designed our agents to be able to win by large score differences. To this aim,
we designed a parametric family of value functions $\nu=\nu_{\lambda}(s)$,
$\lambda\in[0,1]$, as the average of $\sigma(x,\alpha,\beta)$
for $x$  ranging from $\bar k$ to a level of bonus/malus points $\bar x_\lambda$ that would make the game closer to be even: in other words, for $\lambda>0$, $\nu_{\lambda}(s)$ under- or over-estimates the probability of victory, according to whether the player is winning or losing.

\section{Proof of concept: 7$\times$7 SAI}

\subsection{Scaling down Go complexity}

Scaling the Go board from size $n$ to size $\rho n$ with $\rho<1$
yields several advantages:
\begin{itemize}
	\item Average number of legal moves at each position scales by
	$\rho^2$.
	\item Average length of a game scales by $\rho^2$.
	\item The number of visits in the UC tree that would result in a
	similar understanding of the total game, scales at an unclear
	rate, nevertheless one may naively infer from the above two, that
	it may scale by about $\rho^4$.
	\item The number of resconv layers in the ANN tower scales by $\rho$.
	\item The fully connected layers in the ANN are also much smaller,
	even if it is more complicated to estimate the speed contribution.
\end{itemize}
All in all it is reasonable that the total speed improvement for
self-play games is of the order of $\rho^9$ at least.

Since the expected time to train 19$\times$19 Go on reasonable
hardware has been estimated to be in the order of several hundred
years, we anticipated that for 7$\times$7 Go this time should be in
the order of weeks.
In fact, with a small cluster of 3 personal computers with average
GPUs we were able to complete most runs of training in less than a
week each.
We always used networks with 3 residual convolutional layers of 128
filters, the other details being the same as Leela Zero.
The number of visits corresponding to the standard value of 3200 used
on the regular Go board would scale to about 60 for 7$\times$7. We
initially experimented with 40, 100 and 250 visits and then went with
the latter, which we found to be much better.
The Dirichlet noise $\alpha$ parameter has to be scaled with the size
of the board, according to \cite{AlphaZero} and we did so, testing
with the (nonscaled) values of $0.02$, $0.03$ and $0.045$.
The number of games on which the training is performed was assumed to
be quite smaller that the standard 250k window used at size 19, and
after some experimenting we observed that values between 8k and 60k
generally give good results.

\subsection{Neural network structure}

As explained in Section~\ref{s:model_nn}, Leela Zero's neural network
provides for each position two outputs: policy and winrate. SAI's neural network should provide for each position three outputs:
the policy as before and the two parameters $\alpha$ and $\beta$ of a
sigmoid function which would allow to estimate the winrate for
different komi values with a single computation of the net.
It is unclear whether the komi itself should be provided as an input
of the neural network: it may help the policy adapt to the
situation, but could also make the other two parameters
unreliable\footnote{As will be explained soon, the training is done at
  the level of winrate, so in principle, knowing the komi, the net
  could train $\alpha$ and $\beta$ to any of the infinite pairs that,
  with that komi,  give the right winrate.}. For the initial experiments
the komi will not be provided as an input to the net.

With the above premises, the first structure we propose for the
network is very similar to Leela Zero's one, with the value head
substituted by two identical copies of itself devoted to the
parameters $\alpha$ and $\beta^*$. The latter is then mapped to
$\beta$ by equation $\beta_s=c \exp(\beta^*_s)$.
The exponential transform imposes the natural condition that $\beta$
is always positive. The constant $c$ is clearly redundant when the net
is fully trained, but the first numerical experiments show that it may
be useful to tune the training process at the very beginning, when the
net weights are almost random, because otherwise $\beta$ would be close to 1, which is much too large for random play, yielding
training problems.
The two outputs were trained with the usual $l^2$ loss function
but with the value $v_s$ substituted with $\rho_s(0)
=\sigma(\bar k_s,\alpha_s,\beta_s)$.

We used two structures of network, \emph{type V} and \emph{type Y}, which are described in detail in~\cite{SAIpreprint}.

\subsection{Branching from intermediate positions}

To train the network we included the komi value into the training data
used by SAI. The training is then performed the same way as for
Leela Zero, with the loss function given by the sum of regularization
term, cross entropy for the policy and $l^2$ norm for the winning
rate.

The winning rate is computed with the sigmoid function given by
equations~\eqref{e:rho_def} and~\eqref{e:sigma_def}, in particular we
set $v(s)=\rho_s(0)$ and backpropagate gradients through these
functions.

To train the neural network it is clearly necessary to have different
komi values in the data set. It would be best to have \emph{very}
different komi values, but when the agent starts playing well enough,
only few values around the correct komi\footnote{The correct komi for
  7$\times$7 Go is known to be 9, in that with that value both players
  can obtain a draw. Since we didn't want to deal with draws, for
  7$\times$7 Leela Zero we chose a $9.5$ komi, thus giving victory to
  white in case of a perfect play. In fact we noticed that with a komi
  of $7.5$ or $8.5$ (equivalent by chinese scoring) the final level of
  play of the agents didn't seem to be as subtle as it appears to be
  for the $9.5$ komi.}  make the games meaningful.

To adapt the komi values range to the ability of the current network,
when the server assign a self-play match to a client, it chooses a
komi value randomly generated with distribution given by the sigmoid
itself. Formally,
\begin{equation}
K=0.5+\lfloor\rho_s^{-1}(U)\rfloor
\end{equation}
where $\rho_s(x)=\sigma(x,\alpha_s,\beta_s)$, $s$ is the initial empty
board state, $\alpha_s$ and $\beta_s$ are the computed values with
current network and $U\sim\unif(0,1)$, thus giving to $K$ an
approximate logistic distribution.

As the learning goes on, we expect $\alpha_s$ to converge to the
correct value of 9, and $\beta_s$ to increase, narrowing the range of
generated komi values.

To deal with this problem we implemented the possibility for the
server to assign self-play games starting from any intermediate
position.

After a standard game is finished, the server looks to each of the
game's positions and from each one may branch a new game
(independently and with small probability). The branched game starts
at that position with a komi value that is considered even by the
network. Formally,
\[
k'=0.5+\lfloor\pm\alpha_s\rfloor
\]
where $s$ is the branching position and $\pm\alpha_s$ is the value of
$\alpha$ at position $s$, as computed by the current network, with the
sign changed if the current player was white.

The branched game is then played until it finishes and then all its
positions starting from $s$ are stored in the training data, with komi
$k'$ and the correct information on the winner of the branch.

This procedure should produce branches of positions with unbalanced
situations and values for the komi that are natural to the situation
but nevertheless range on a wide interval of values.

\subsection{Sensible agent}\label{sensibleagent}

When SAI plays, it can estimate the winrate for all values
of the komi with a single computation of the neural network. In fact,
getting $\alpha$ and $\beta$ it knows the sigmoid function that gives
the probability of winning with different values of the komi for the
current position.

We propose the generalization of the original agent of Leela Zero
as introduced in Section~\ref{s:agent_model}. Here we give further details.

The agent behaviour is parametrized by a real number $\lambda$ which
will be usually chosen in $[0,1]$. 

To describe rigorously the agent, we need to introduce some
more mathematical notation.

\paragraph{Games, moves, trees.}

Let $\mathcal G$ be the set of all legal game states, with
$\varnothing\in\mathcal G$ denoting the empty board starting state.

For every $s\in\mathcal G$, let $\mathcal A_s$ the set of legal moves
at state $s$ and for every $a\in\mathcal A_s$, let $s_a\in\mathcal G$
denote the game state reached from $s$ by performing move $a$. This
clearly induces a directed graph structure on $\mathcal G$ with no
directed cycles (which are not legal because of superko rule) and with
root $\varnothing$. This graph can be uplifted to a rooted tree by
taking multiple copies of the states which can be reached from the
root by more than one path. From now on we will identify $\mathcal G$
with this rooted tree and denote by $\rightarrow$ the edge relation
going away from the root.

For all $s\neq\varnothing$ let $\bar s$ denote the unique state such
that $\bar s\rightarrow s$.

For all $s\in\mathcal G$, let
$\mathcal R_s=\{r\in\mathcal G:s\rightarrow r\}$ denote the set of
states reachable from $s$ by a single move. We will identify
$\mathcal A_s$ with $\mathcal R_s$ from now on.

For any subtree $T\subset\mathcal G$, let $|T|$ denote its size
(number of nodes) and for all $s\in T$, let $T_s$ denote the subtree
of $T$ rooted at $s$.

\paragraph{Values, preferences and playouts.}

Suppose that we are given three maps $P$, $u$ and $v$, with the
properties described below.
\begin{itemize}
\item The \emph{policy} $P$, defined on $\mathcal G$ with values in $[0,1]$
  and such that
  \[
    \sum_{r\in\mathcal R_s}P(r)=1
    ,\qquad s\in\mathcal G.
  \]
  This map represents a measure of goodness of the possible moves.
\item The \emph{value} $v$, defined on
  $\{(s,r):s\in\mathcal G, r\in\mathcal G_s\}$ with values in $[0,1]$,
  which represents a rough estimate of the winrate at a
  future state $r$. The estimate is from the point of view of
  whichever player is next to play at state $s$.
\item The \emph{first play urgency} $u$, defined for all pairs $(s,T)$
  such that $s\in\mathcal G$ and $T\subset\mathcal G$ with values in
  $[0,1]$. This represents an ``uninformed'', flat, winning rate
  estimate of all states in $\mathcal R_s\setminus T$, i.e.~actions
  which were not yet visited. It may depend on the set $T$ of visited
  states.
\end{itemize}

Then for any non-empty subtree $T$ and node $s$ not necessarily inside $T$ we
can define the \emph{evaluation} of $s$ over $T$, as
\[
  Q_T(s):=
  \begin{cases}
    u(\bar s,T) & \text{if }s\notin T \\[1ex]
    \displaystyle\frac1{|T_s|}\sum_{r\in T_s}v(\bar s,r) &  \text{if }s\in T
  \end{cases}
\]
It should be noted here that the two proposed choices for $u$ are the
following:
\begin{gather}
  u(s,T)\equiv0.5 \tag{AlphaGo Zero} \\
  u(s,T)
  =v(s,s)
  -C_{\text{fpu}}\sqrt{\sum_{r\in \mathcal R_s\cap T}P(r)} \tag{Leela Zero}
\end{gather}  
We can then define the \emph{UC urgency} of $s$ over $T$, as
\[
  U_T(s)
  :=Q_T(s)+C_{\text{puct}}\sqrt{|T_{\bar s}|-1}\frac{P(s)}{1+|T_s|}
\]

Finally, the \emph{playout} over $T$, starting from $s\in T$ is
defined as the unique path on the tree which starts from $s$ and at
every node $r$ chooses the node $t\in\mathcal R_r$ that maximizes
$U_T(t)$.

\paragraph{Definition of $v$.}

In the case of Leela Zero, the value function $v(s,r)$ depends on $s$
only through parity: let $\hat v_r$ be the estimate of the winning
rate of current player at $r$, i.e.~the output of the value head of
the neural network, passed through an hyperbolic tangent and rescaled
in $(0,1)$. Then
\[
v(s,r):=\begin{cases}
  \hat v_r   & s,r\text{ with same current player} \\
  1-\hat v_r & s,r\text{ with different current player}
\end{cases}
\]

In the case of SAI, the neural network provides the sigmoid's
parameters estimates $\hat\alpha_r$ and $\hat\beta_r$ for the state
$r$. These allow to compute the estimate $\hat\rho_r$ of the winning
probability for the current player at all komi values.
\[
  \hat\rho_r(x)
  :=\sigma(\hat\beta_r(\hat\alpha_r+\bar k_r+x))
\]
Here $\bar k_r$ is the official komi value from the perspective of the
current player, at state $r$,
\[
\bar k_s:=\begin{cases}
k & \text{if at $s$ the current player is white} \\
-k & \text{if at $s$ the current player is black},
\end{cases}
\]
the \emph{komi correction} $x$ is a real variable that allows to fake
an arbitrary virtual komi value, and $\sigma$ is the standard logistic
sigmoid,
\[
  \sigma(x)
  :=\frac1{1+e^{-x}}
  =\frac12+\frac12\tanh\biggl(\frac x2\biggr).
\]
Then if we want SAI to simulate the playing style of Leela Zero,
though with its own understanding of the game situations, we can
simply let
\[
v(s,r):=\begin{cases}
  \hat\rho_r(0)   & s,r\text{ with same current player} \\
  1-\hat\rho_r(0) & s,r\text{ with different current player.}
\end{cases}
\]
On the other hand, if we want SAI to play ``sensibly'', we may use
values of $x$ for which $\hat\rho_r(x)$ is away from 0 and from 1, so
that it can better distinguish the consequences of its choices, as
they reflect more in the winrate. This means to give the agent a
positive virtual komi correction if it is behind and a negative
virtual komi correction if it is ahead.

One way this can be done in a robust way, is to compute the average of
the expected winrate at the future state $r$ over a range of komi
correction values that depends on the current state $s$: an interval
of positive numbers if the net believes that $s$ is losing and
negative if winning.

By deciding the interval at $s$, we are avoiding situations like when
the current player is winning at $s$, it explores a sequence of future
moves with a blunder, so that it is losing at $r$, and then evaluates
the winrate at $r$ giving itself a bonus which will then mitigate the
penalization.

In fact, in this way blunders done when ahead are penalized more than
before in the exploration, which seems a good feature.

Formally, we introduce the symbol $\mu_r(y)$ to denote the average of
$\hat\rho_r$ over the interval $[0,y]$ or $[y,0]$,
\begin{equation}\label{e:mu_def}
  \mu_r(y):=\begin{cases}
    \hat\rho_r(0) & y=0\\[1ex]
    \displaystyle\frac1y\int_0^y\hat\rho_r(x)dx & y\neq0.
  \end{cases}
\end{equation}
Let the \emph{common sense parameter} $\lambda$ be a real parameter, usually in
$[0,1]$, and let $\pi_\lambda$ be
\[
\pi_\lambda:=(1-\lambda)\hat\rho_s(0)+\lambda\frac12
\]
so that $\pi_0=\hat\rho_s(0)$, $\pi_1=\frac12$ and $\pi_\lambda$ a
convex combination of the two for $\lambda\in[0,1]$. We introduce the
extremum of the komi correction interval as the reverse image of
$\pi_\lambda$,
\begin{gather*}
  \bar x_{s,\lambda}
  :=\hat\rho_s^{-1}(\pi_\lambda)
\end{gather*}

Then for a version of SAI which plays with parameter $\lambda$, we
let the value be defined by,
\[
v(s,r):=\begin{cases}
  \mu_r(\bar x_{s,\lambda})   & s,r\text{ with same current player} \\
  1-\mu_r(\bar x_{s,\lambda}) & s,r\text{ with different current player.}
\end{cases}
\]
Hence the value is computed at state $r$ but the range of the average
is decided at state $s$.

\begin{rem}
  We bring to the attention of the reader that a simple rescaling
  shows that the quantity $\mu_r(\bar x_{r,\lambda})$ would be
  somewhat less useful, because it depends on $\hat\alpha_r$ and
  $\hat\beta_r$ only through $\hat\rho_r(0)$.
\end{rem}

\begin{rem}
  As shown in~\cite{SAIpreprint}, the integral in equation~\eqref{e:mu_def} can be
  computed analytically and easily implemented in the software.

\begin{multline*}
  \mu_r(y)
  =\frac12+\frac{b-a}{2y}-\frac1{\hat\beta_ry}\log\sigma\bigl(\hat\beta_rb\bigr)+\frac1{\hat\beta_ry}\log\sigma\bigl(\hat\beta_ra\bigr)
\end{multline*}
where $a:=|\hat\alpha_r+\bar k_r|$ and $b:=|\hat\alpha_r+\bar k_r+y|$.
\end{rem}

\paragraph{Tree construction and move choice.}

Suppose we are at state $t\in\mathcal G$ and the agent has to choose a
move in $\mathcal A_t$. This will be done by defining a suitable
\emph{decision} subtree $\mathcal T$ of $\mathcal G$, rooted at $t$,
and then choosing the move $s$ randomly inside  $\mathcal R_t$ with
probabilities proportional to
\[
  \exp(C_{\text{temp}}^{-1}|\mathcal T_s|)
  ,\qquad s\in\mathcal R_t
\]
where $C_{\text{temp}}$ is the Gibbs temperature which is defaulted to
1 for the first moves of self-play games and to 0 (meaning that the
move with highest $|\mathcal T_s|$ is chosen) for other moves and for
match games.

The decision tree $\mathcal T$ is defined by an iterative
procedure. In fact we define a sequence of trees
$\{t\}=:T^{(1)}\subset T^{(2)}\subset \dots$ and stop the procedure by letting
$\mathcal T:=T^{(N)}$ for some $N$ (usually the number of \emph{visits} or
when the thinking time is up).

The trees in the sequence are all rooted at $t$ and satisfy
$|T^{(n)}|=n$ for all $n$, so each one adds just one node to the previous
one:
\[
T^{(n)}=T^{(n-1)}\cup\{t_n\}  
\]

The new node $t_n$ is defined as the first node outside $T^{(n-1)}$
reached by the playout over $T^{(n-1)}$ starting from $s$.

\subsection{Measuring playing strength}

To provide a benchmark for the developement of SAI, we adapted
Leela Zero to 7$\times$7 Go board and performed several runs of
training from purely random play to a level at which further
improvement wasn't expected. More details on this step can be found in~\cite
{SAIpreprint}. A sample of 7$\times$7 Leela Zero nets formed the panel used in the evaluation phase of the SAI runs.

When doing experiments with training runs of Leela Zero, we produce
many networks, which had to be tested to measure their playing strength,
so that we can assess the performance and efficiency of each run.

The simple usual way to do so is to estimate an Elo/GOR score for each
network\footnote{In fact the neural network, is just one of many
	components of the playing software, which depends also on several
	other important choices, such as the number of visits, fpu policies
	and all the other parameters. Rigorously the strength should be
	defined for the playing \emph{agent} (each software implementation
	of Leela Zero), but to ease the language and the exposition, we will
	speak of the strength of the \emph{network}, meaning that the other
	parameters were fixed at some value for all matches.}. The idea
which defines this number is that if $s_1$ and $s_2$ are the scores of
two nets, then the probability that the first one wins against the
second one in a single match is
\[
\frac1{1+e^{(s_2-s_1)/c}}
\]
so that $s_1-s_2$ is, apart from a scaling coefficient $c$
(traditionally set to 400), the log-odds-ratio of winning.

This model is so simple that is actually unsuitable to deal with the
complexity of Go and Go playing ability. In fact in several runs of Leela Zero 7$\times$7 we observed that each
training phase would produce at least one network which solidly won
over the previous best, and was thus promoted to new best. This
process would continue forever, or at least as long as we dared keep
the run going, even if from some point on, the observed playing style
was not evolving anymore. When some match was tried between
non-consecutive networks, we saw that the strength inequality was not
transitive, in that it was easy to find cycles of 3 or more networks
that regularly beat each other in a directed circle. Even with very
strong margins.

We even tried to measure the playing strength in a more refined way,
by performing round-robin tournaments between nets and then estimating
Elo score by maximum likelihood methods. This is much heavier to
perform and still showed poor improvement in predicting match outcomes.

It must be noted that this appears to be an interesting research
problem in its own. The availability of many artificial playing agents
with different styles, strengths and weaknesses will open new
possibilities in collecting data and experimenting in this field.

\begin{rem}
	It appears that this problem is mainly due to the peculiarity of
	7$\times$7 Go and only relevant to it.
	
	In the official 19$\times$19 Leela Zero project the Elo estimation
	is done with respect to previous best agent only and it is known
	that there is some Elo inflation, but tests against a fixed set of
	other opponents or against further past networks have shown that
	real playing strength does improve.
\end{rem}  


A different approach which is both robust and refined and is easy to
generalize is to use a panel of networks to evaluate the strength of
each new candidate.

We chose 15 networks of different strength from the first 5 runs of
Leela Zero 7$\times$7. Each network to be evaluated is opposed to each
of these in a 100 games match. The result is then a vector of 15
sample winning rates, which contains useful multivariate information
on the playing style, strengths and weaknesses of the tested net.

To summarize this information in one rough  scalar score number,
we used principal component
analysis (PCA). We performed covariance PCA once for all the match
results of the first few hundreds of good networks, determined the
principal factor and used its components as weights%
\footnote{%
	Due to the asymmetric choice of the default komi at 9.5, which is favorable to white, the experiment showed that after a run reached almost perfect play as white, its playing style as black would typically get randomly worse, as the learning frantically tried to find any way for black to trick white and get an impossible win. Hence we had to use only the results of the nets playing as white to evaluate performaces. We recall that, by the properties of PCA, the principal factor will find the maximum distinguisher between results against the panel networks. In the 15-dimensions space of results this will be the direction in which the networks analysed are more different from one another. The coefficients of this factor resulted to be all positive numbers ranging from 0.001 for very weak networks, to 0.62 for the strongest one, thus confirming that these weights represent a reasonable measure of strength.%
}. Hence the score of a network is the principal component of its PCA
decomposition. This value, which we call \emph{panel evaluation}, correlates well
with the maximum likelihood estimation of Elo by round-robin matches,
but is much easier and quicker to compute.

\section{Results}

\subsubsection{Obtaining a strong SAI}

\begin{table}[ht]
	\footnotesize
	\def\htab{.25cm}
	\setlength{\tabcolsep}{3pt} 
	\renewcommand{\arraystretch}{1.2} 
	\begin{tabular}{l*{2}{c} *{5}{c} cr}\centering 
		&  \multicolumn{2}{c}{\bf  Network } &  \multicolumn{5}{c}{\bf Parameters during self-play} &  \multicolumn{2}{c}{\bf Results }\\ 
		\cline{2-3} \cline{4-8} \cline{9-10}
		\multicolumn{1}{c}{\bf Run} &  {\bf Type } & {\bf IP}  & {\bf  MV } & {\bf  FPU } & {\bf ST } & {\bf RT } & {\bf $\lambda$ }   & {\bf PE} & {\bf TTP }\\ 
		\hline
		SAI 9  & V & 18& 250 & AGZ & 1.0 & 1.0  & 0.5  & 2.34848 & 1838 \\ 
		SAI 15 & Y & 18& 250 & AGZ & 2.0 & 0.5  & 0.0  & 1.94057 & 1194  \\
		SAI 20 & Y & 17& 250 & LZ  & 1.5 & 0.8  & 0.5  & 2.38155 & 463  \\
		SAI 22 & Y & 17& 502 & LZ  & 1.5 & 0.8  & 0.5  & 2.39176 & 2545 \\
		SAI 23 & Y & 17& 250 & LZ  & 1.0 & 1.0  & 0.5  & 2.35204 & 465\\
		SAI 24 & Y & 17& 250 & LZ  & 1.0 & 1.0  & 0.0  & 2.38220 & 528\\ 
		\hline
	\end{tabular}
	\caption{Description of a representative sample of SAI runs.  \textbf{Type}: shape of the network. \textbf{IP}: number of input planes. \textbf{MV}: max visits. \textbf{FPU}: first playing urgency, with values AGZ (AlphaGo Zero) and LZ (Leela Zero). \textbf{ST}: softmax temperature. \textbf{RT}: random temperature. \textbf{$\lambda$}: parameter of the agent.  \textbf{PE}: panel evaluation of the 3rd best net of the run. \textbf{TTP}: time to plateau, in millions of nodes.}
	\label{t:evolutionSAI}
\end{table}

\begin{figure}[ht]
  \begin{center}
    \includegraphics[width=.42\textwidth]{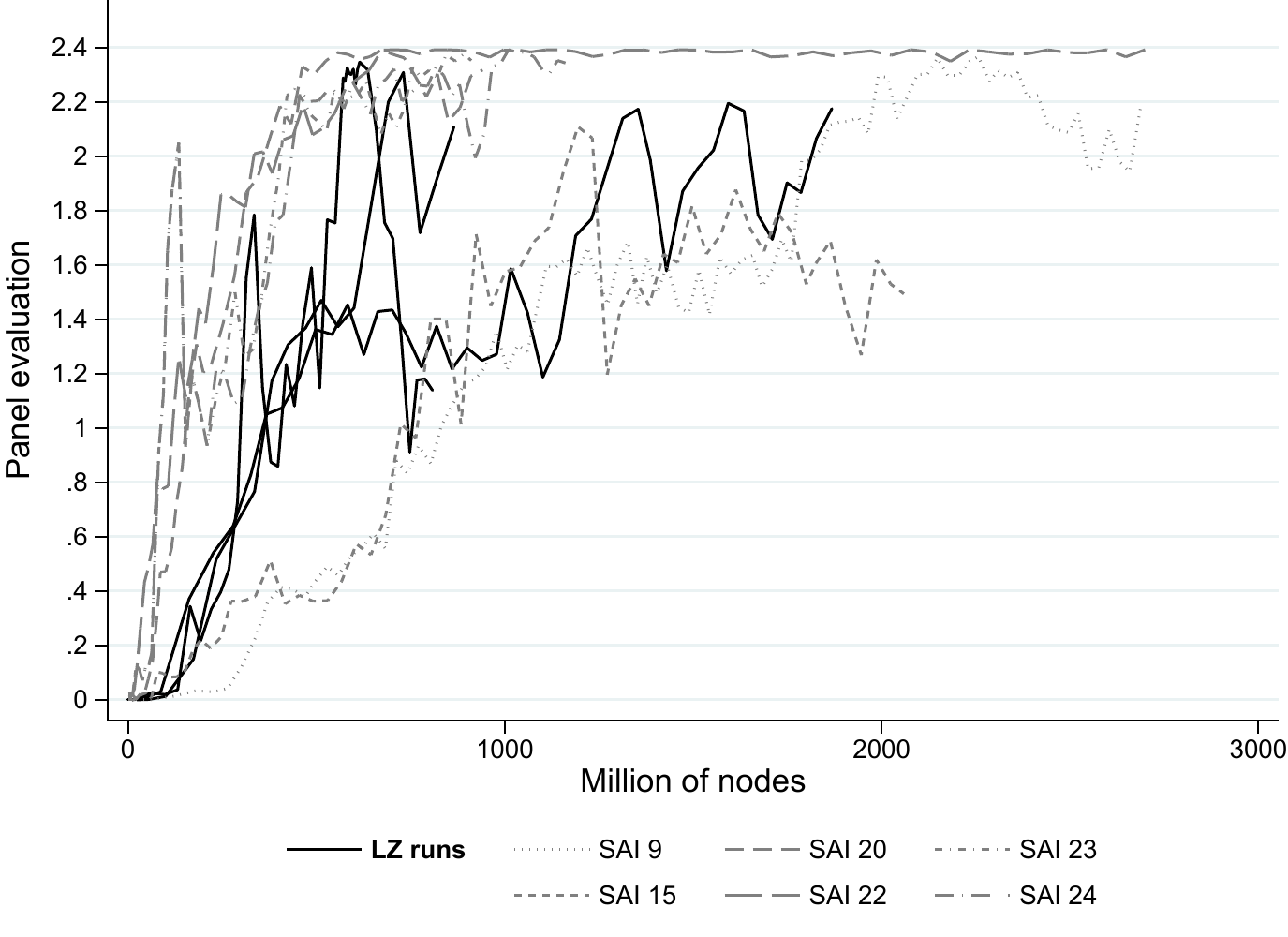}
  \end{center}
\figbotadj
  \caption{Panel evaluation, as a function of the millions nodes computed. Best three runs of Leela Zero reported as a landmark.}
  \label{f:sai_main}
\end{figure}

The first runs of SAI failed to reach the performance of the reference $7\times7$ Leela Zero runs. A turning point was the 9th run, when we simplified the formula for the branching probability and assigned constant probability of branching
$C_\text{branch}=0.025$ for all states, thus giving higher chance of
branching in balanced situation. This resulted in a steady and important improvement. In Table~\ref{t:evolutionSAI} we summarized the characteristics of the most representative runs we ran after the 9th, together with their performance, measured as the panel evaluation of the 3rd best net of the run, and efficiency, measured in terms of time to reach the plateau level. In Figure~\ref{f:sai_main} we represented the evolution of the  performance of the same runs, across millions of nodes. Run 15 was the lowest point, showing that increasing the softmax temperature too much, while decreasing the random temperature, produced negative results. After run 15th we also settled for the Leela Zero form of first playing urgency, as opposed to AlphaGo Zero's. Run 20th had the best balance between performance and efficiency. Increasing the maximum number of visits in run 22nd resulted in a severe loss of efficiency, not adequately compensated by a gain in performance. In runs 23rd and 24th the two temperature parameters were slightly modified again, and  $\lambda$ was set to 0 and 0.5 respectively, without significant gains.
\figtopadj

\subsubsection{Evaluation of positions by Leela Zero and by SAI}

\begin{figure}[ht]
  \begin{center}
    \includegraphics[width=.48\textwidth]{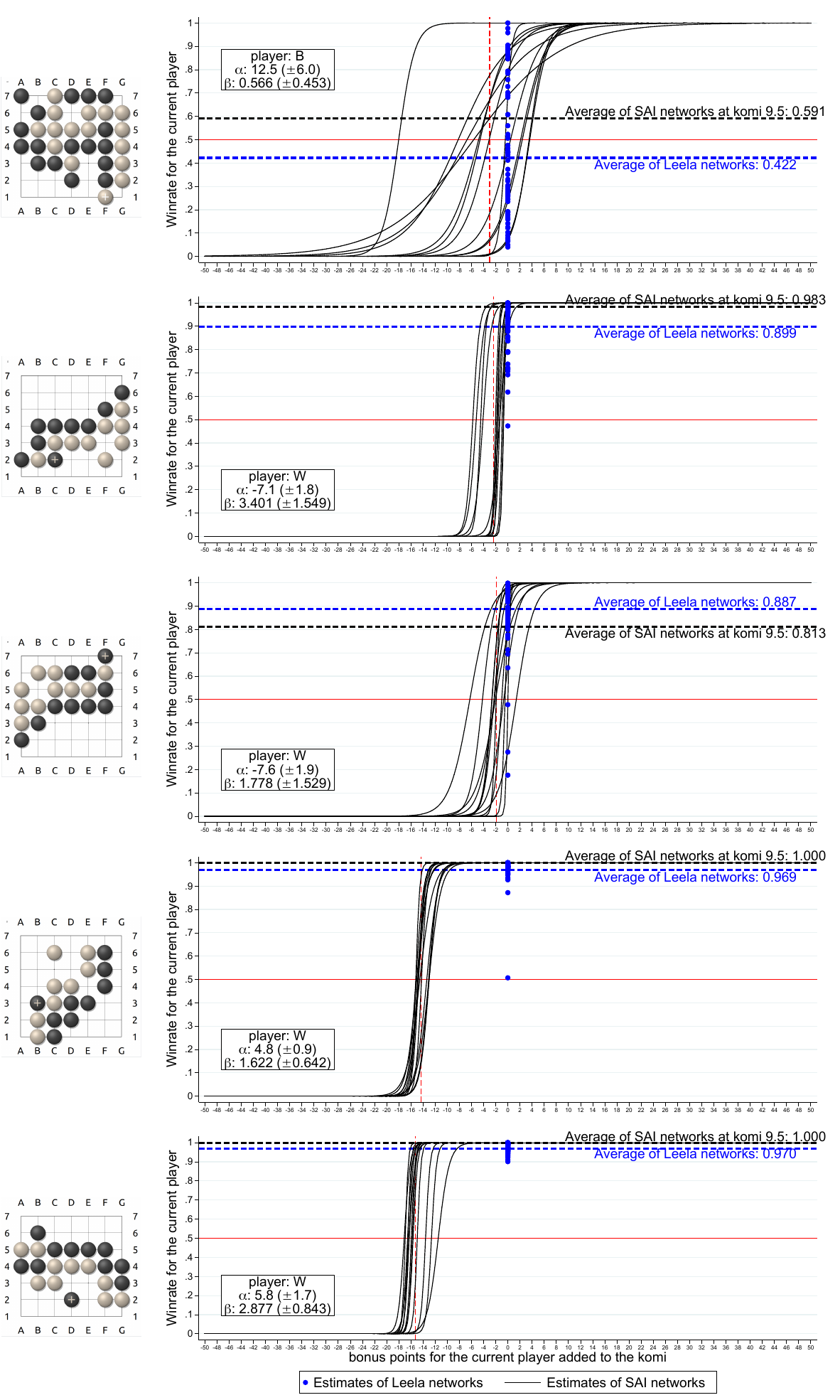}\\[2ex]
  \end{center}
\figbotadj
  \caption{Evaluation of five positions by a sample of strong $7\times7$ Leela Zero and SAI nets.} \label{f:evaluation5pos}
\end{figure}

To illustrate the ability of SAI to understand the winrate
in a more complex fashion, we chose 5 meaningful positions which are
shown in Figure~\ref{f:evaluation5pos}. The first 3 positions were chosen 
from a sample of games as the most frequent which offered two different winning moves, one with higher score than the other. The 4th and 5th positions were created \emph{ad hoc} as positions where the victory is granted for the current player, but two different moves give a different score.
\figtopadj
For each position we plotted SAI's sigmoid evaluations of the winrate
(black curves) and $7\times7$ Leela Zero's point estimates of winrate at standard komi (blue
dots). Every one of these plots shows a sample of 63 $7\times7$ Leela Zero and 13
SAI nets from different runs, chosen among the strongest ones.

It is important
to observe that the distributions of the winrates seem to agree for
the two groups at standard komi, indicating that SAI's estimates have
similar accuracy and precision as $7\times7$ Leela Zero's.

%
%

The SAI nets provide an estimate of the difference of points between the players. The variability that we observe shows that even strong nets do
not have a uniform understanding of single complicated positions. However we can observe that the wider the discrepancies among estimates of $\alpha$, the lower  the estimate of $\beta$, thus showing that the nets are aware that the estimate is unstable. This confirms the robustness of our approach.

We analyse separately each position, using human expertise.
\\
\textbf{Position 1.} Black, the current player, is ahead of 13
points on the board, thus, with komi 9.5, his margin is 3.5
points. However the position is difficult, because there is a \emph{seki}: this is a situation when 
an area of the board provides points (is alive) for both players 
(quite uncommon in our 7$\times$7 games), and may be poorly
interpreted as white dead (black ahead by 49 points on the board) or as
black dead (black ahead by 5 points on the board). In agreement with this
analysis, the sample of SAI nets gives a low and sharp estimate for
$\beta$ with average $0.566$ and standard deviation $0.453$ and a
wild estimate for $\alpha$, with average $12.5$ and standard
deviation $6.0$. The sample of
Leela Zero nets gives winrate estimates which are almost uniformly
distributed in $[0,1]$: many of these nets have an incorrect
understanding of the position and are not aware of this. SAI nets on
the other hand are aware of the high level of uncertainty.\\
\textbf{Position 2.} White, the current player, is behind by 5
points on the board, thus, with komi 9.5, she is winning by 4.5
points. Following the policy, which recognizes a common shape
here, many nets will consider cutting at F6 instead of E5, therefore
losing one point. Accordingly, the estimate of $\alpha$ ranges
approximately from $-5.5$ to $-8.5$ with average $-7.1$ and standard
deviation $1.8$. The sample of $\beta$ has average $3.401$ and
standard deviation $1.549$, thus showing that $\alpha$ is to be
considered precise up to two units.\\
\textbf{Position 3.} Here the situation is very similar to the previous
one: white is behind by 7 points on the board, thus, with komi 9.5,
white is winning by 2.5 points. Following the policy, which
recognizes a common shape here, many nets will consider cutting at
B2 instead than C3, therefore losing one point. Accordingly, the estimate
of $\alpha$ ranges approximately from $-5.5$ to $-9.5$ with average
$-7.6$ and standard deviation $1.9$. The sample of $\beta$ has
average $1.778$ and standard deviation $1.529$, thus showing that
$\alpha$ is to be considered precise up to two units.\\
\textbf{Position 4.} White, the current player, 
is ahead by 5 points on the board, thus, with komi she is winning by a larger margin of 14.5 points.
Following the policy, white is facing the choice between B4 and A3, capturing the single black stone. There is a slight strategic difference between B4 and A3: A3 is better in case a \emph{ko}\/ fight emerges.
Accordingly, we found a sharp estimate for $\alpha$ ranging from 4 to $5.5$, with average $4.8$ and standard deviation $0.8$. The sample of $\beta$ has average $1.622$ and standard deviation $0.642$.\\
\textbf{Position 5.} White, the current player, is ahead by 5 points on the board, thus, with komi, she is winning by a larger margin of $14.5$ points.
The position is particularly easy to understand: white will win with every possible move on the board, including the pass; although
only the move A3 gives white the largest possible victory. Accordingly, the estimate of $\alpha$ range from 4 to $7.5$ with average $5.8$ and standard deviation $1.7$. The sample of $\beta$ has average $2.877$ and standard deviation $0.843$.

\subsubsection{Experimenting different agents for SAI}

\begin{figure}[ht]
  \begin{center}
    \includegraphics[width=.46\textwidth]{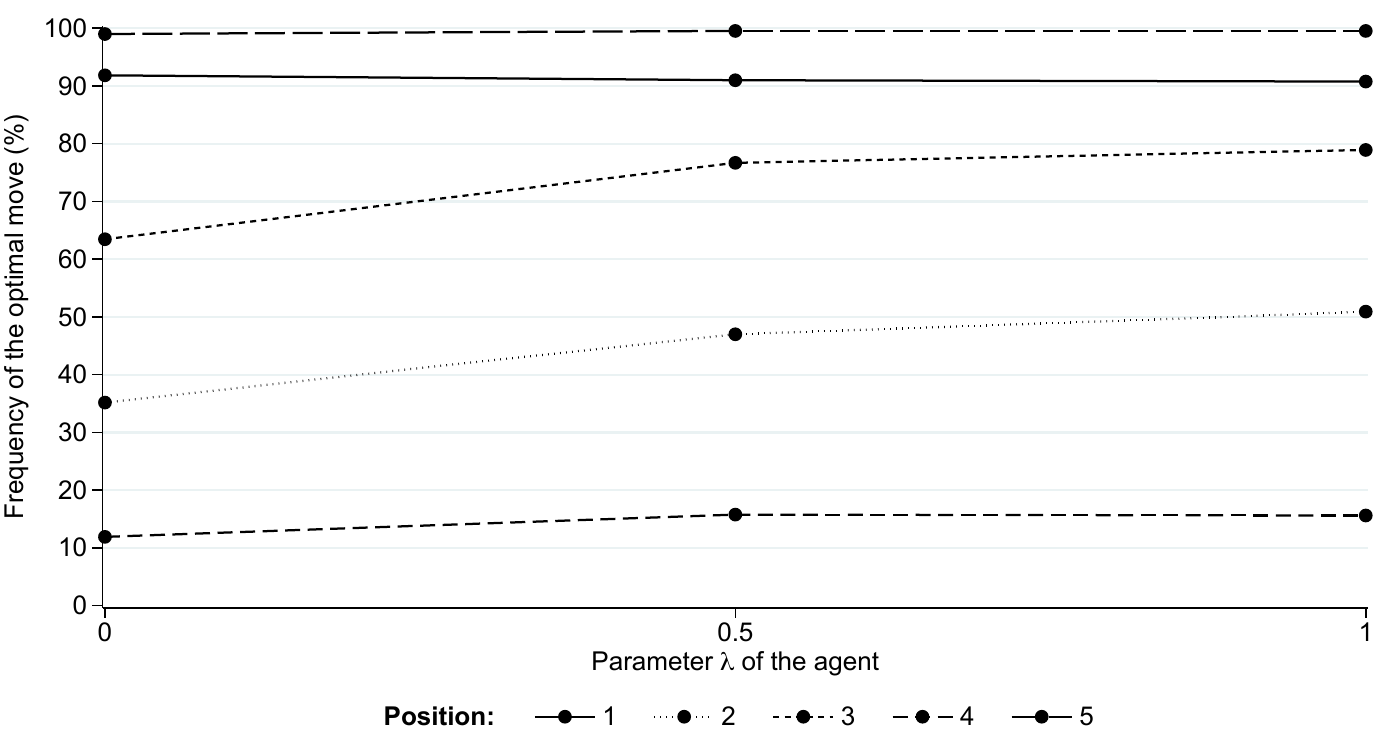}
  \end{center}
\figbotadj
  \caption{Probability that a net chooses the optimal move from each of the five positions, for increasing values of the $\lambda$ parameter.} \label{f:experimentlambda}
\end{figure}

Finally, we experimented on how the parameter $\lambda$ of
the agent affects the preference of the next move, from positions where at least two winning moves are available. This was
done using the 5 positions shown in Figure~\ref{f:evaluation5pos} and asking to the same 13 SAI nets to choose the next move. The parameter $\lambda$ was set to 0, 0.5 and 1, 1000 times each. In Figure~\ref{f:experimentlambda} the results are represented. In position 1 and 5 the optimal move was chosen more than 90\% of times for $\lambda=0$ already, and incresing $\lambda$ did not affet the choice. In the other 3 positions  increasing $\lambda$ improved the choice of the optimal move, as expected. 
\figtopadj

\section{Conclusions}

We introduced SAI, a reinforcement learning solution for playing Go
which generalizes the previous models to multiple komi. The winrate as
function of komi is estimated by a two-parameters family of sigmoid
curves.  We performed several complete training runs on the simplified
7$\times$7 goban, exploring parameters and settings, and proving that
it is more difficult, but possible, to effectively train the net to
learn two continuous parameters in spite of the fact that the match
outcome is a single binary value (win/lose). The generation of a
suitable ensemble of game branches with adjusted komi appears to be a
key point to this end.

The estimates of the winrate of our nets at standard komi
are compatible with those of Leela Zero, but at the same time SAI's
winrate curves provide a deeper understanding of the game
situation. As a side effect, a good estimate of the final point
difference between players can also be deduced from the winrate
curves.

In principle the winrate curve estimation allows to design sensible
agents that aim to win by larger margins of points against weaker
opponents, or that can play with handicap in points and/or stones. We
propose such an agent, parametrized by a \emph{common sense} parameter
$\lambda\geq0$. When $\lambda=0$ the agent behaves like previous
models and only tries to win. (We could obtain nets able to play at
almost perfect level at $\lambda=0$.)

With $\lambda>0$ the agent is designed to try to win by a high margin of points, 
while still focusing on winning. Due to the
limitations of the 7$\times$7 goban, it was not possible to assess
whether our model could really target higher margins of victory against weak
opponents, but we showed the expected effect of different
values of $\lambda$ on the move selection.

We posit that it should be feasible to implement SAI in the 9$\times$9
and full 19$\times$19 board. Albeit the configuration of the learning
pipeline presents more difficulties than standard Leela Zero and the
training could be longer, the experiments performed on the 7$\times$7
board should be useful to make the right choices and develop some
understanding of the possible unwanted behaviours in order to avoid
them.

The development of a 19$\times$19 board version of SAI with a
distributed effort could produce a software tool able to provide a
deeper understanding of the potential of each position, to target high
margins of victory and play with handicap, thus providing an opponent
for human players which never plays sub-optimal moves, and ultimately
progressing towards the optimal game.

\bibliographystyle{IEEEtranN}
\bibliography{sai}

\end{document}